\documentclass[
]{ceurart}

\usepackage{listings}
\usepackage{tikz}
\usetikzlibrary{positioning} %
\usetikzlibrary{arrows.meta}
\usetikzlibrary{decorations.pathreplacing}
\usepackage[noabbrev]{cleveref}

\usepackage{colortbl}
\definecolor{mygreen}{RGB}{40,140,40}

\usepackage{adjustbox}

\usepackage{siunitx}
\let\model\textsc
\newcommand{\copybs}[0]{\model{copy}}

\newcommand{\witbest}[0]{$\witletterM{}_{\text{\textsc{best}}}$}
\newcommand{\witsqrt}[0]{$\witletterM{}=0.5$}

\newcommand{\witemp}[0]{corpus-frequency temperature}
\newcommand{\Witemp}[0]{Corpus-frequency temperature}

\newcommand{\param}[1]{\texttt{#1}}

\newcommand{\adrop}[0]{\param{attn drop}}
\newcommand{\ddrop}[0]{\param{drop}}
\newcommand{\fdrop}[0]{\param{activation drop}}
\newcommand{\ldrop}[0]{\param{layer drop}}
\newcommand{\head}[0]{\param{attn heads}}
\newcommand{\clipnorm}[0]{\param{grad-clip-norm}}
\newcommand{\lnorm}[0]{\param{L2-norm}}
\newcommand{\layers}[0]{\param{layers}}
\newcommand{\layerdim}[0]{\param{layer dim}}
\newcommand{\ffnndim}[0]{\param{ffnn dim}}
\newcommand{\bs}[0]{\param{batch size}}
\newcommand{\lr}[0]{\param{learning rate}}
\newcommand{\maxeps}[0]{\param{max epochs}}
\newcommand{\ckptsel}[0]{\param{checkpoint sel}}
\newcommand{\decay}[0]{\param{decay}}

\newcommand{\wacc}[0]{token accuracy}

\newcommand{\uacc}[0]{type accuracy}

\newcommand{\wit}[0]{frequency-aware training}

\newcommand{\sig}[0]{SIGMORPHON}
\newcommand{\tr}[0]{Transformer}

\newcommand{\um}[0]{UniMorph}
\newcommand{\ltf}[0]{lemma-tag-form}

\newcommand{\en}[0]{English}
\newcommand{\es}[0]{Spanish}

\newcommand{\eus}[0]{Basque}

\newcommand{\ie}[0]{Indo-European}

\newcommand{\cs}[0]{Czech}

\newcommand{\br}[0]{Breton}

\newcommand{\writtenLatinScript}[0]{written in the Latin script}

\newcommand{\witletter}[0]{$\tau$}

\newcommand{\witletterM}[0]{\tau}

\newcommand{\sweights}[0]{sample weights}
\newcommand{\sweight}[0]{sample weight}

\newcommand{\citepSTs}[0]{\citep{st16-cotterell-etal-2016-sigmorphon,st17-cotterell-etal-2017-conll,st18-cotterell-etal-2018-conll,st20-vylomova-etal-2020-sigmorphon,st21-pimentel-ryskina-etal-2021-sigmorphon,st22-kodner-etal-2022-sigmorphon,st23-goldman-etal-2023-sigmorphon}}

\renewcommand{\cref}[1]{\Cref{#1}}

\usepackage[absolute]{textpos}

\begin{document}
\begin{textblock}{16}(0,0.1)\centerline{This paper was published in \textbf{Information technologies --- Applications and Theory (ITAT) 2025}}\end{textblock}
\begin{textblock}{16}(0,0.3)\centerline{-- please cite the published version {\small\url{https://www.ics.upjs.sk/~antoni/ceur-ws.org/Vol-0000/paper13.pdf}}.}\end{textblock}
\copyrightyear{2025}
\copyrightclause{Copyright for this paper by its authors.
Use permitted under Creative Commons License Attribution 4.0
International (CC BY 4.0).}
\conference{ITAT'25: Information Technologies -- Applications and Theory, September 26--30, 2025, Telgárt, Slovakia}

\title{Corpus Frequencies in Morphological Inflection: Do They Matter?}

\author[]{Tomáš Sourada}[%
orcid=0009-0003-6792-825X,
email=sourada@ufal.mff.cuni.cz,
]
\cormark[1]

\author[]{Jana Straková}[%
orcid=0000-0003-0075-2408,
email=strakova@ufal.mff.cuni.cz,
]

\address[]{Charles University, Faculty of Mathematics and Physics, Institute of Formal and Applied Linguistics, Prague, Czech Republic}

\cortext[1]{Corresponding author.}
\begin{abstract}
The traditional approach to morphological inflection (the task of modifying a base word (lemma) to express grammatical categories) has been, for decades, to consider lexical entries of lemma-tag-form triples uniformly, lacking any information about their frequency distribution.
However, in production deployment, one might expect the user inputs 
to reflect a real-world distribution of frequencies in natural texts.
With future deployment in mind, we explore the incorporation of corpus frequency information into the task of morphological inflection along three key dimensions during system development:
    (i) for train-dev-test split, we combine a lemma-disjoint approach, which evaluates the model’s generalization capabilities, with a frequency-weighted strategy to better reflect the realistic distribution of items across different frequency bands in training and test sets;
    (ii)~for evaluation, we complement the standard \textit{\uacc{}} (often referred to simply as \textit{accuracy}), which treats all items equally regardless of frequency, with \textit{\wacc{}}, which assigns greater weight to frequent words and better approximates performance on running text;
    (iii) for training data sampling, we introduce a method novel in the context of inflection, \textit{frequency-aware} training, which explicitly incorporates word frequency into the~sampling process.
  We show that frequency-aware training outperforms uniform sampling in 26 out of 43 languages.
\end{abstract}

\begin{keywords}
  Morphological inflection \sep
  Frequency-weighting \sep
  Frequency-weighted split \sep
  Frequency-weighted sampling \sep
  Frequency-weighted evaluation
\end{keywords}

\maketitle

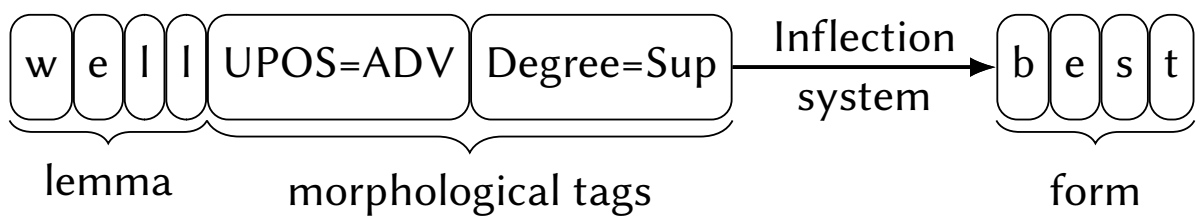
\begin{figure}[h]
\begin{center}
\resizebox{1\hsize}{!}{%
\tikzstyle{rec}=[rectangle, rounded corners=1ex, minimum height = 0.8cm, text height=1.5ex, text depth=0.5ex]

\begin{tikzpicture}[thin, inner sep=0.7ex]

\node[rec, draw] (a) {w};
\node[rec, draw, right=0cm of a] (b) {e};
\node[rec, draw, right=0cm of b] (c) {l};
\node[rec, draw, right=0cm of c] (d) {l};
\node[rec, draw, right=0cm of d] (e) {UPOS=ADV};
\node[rec, draw, right=0cm of e] (f) {Degree=Sup};

\node[rec, draw, right=2cm of f] (g) {b};
\node[rec, draw, right=0cm of g] (h) {e};
\node[rec, draw, right=0cm of h] (i) {s};
\node[rec, draw, right=0cm of i] (j) {t};

\draw [thick, -{Latex[length=2mm]}] (f) -- (g) node[midway, above] {Inflection};
\draw [thick, -{Latex[length=2mm]}] (f) -- (g) node[midway, below] (arr) {system};

\draw[decorate,decoration={brace,amplitude=5pt, mirror}] (a.south west) -- (d.south east) node[midway, below=6pt] {lemma};

\draw[decorate,decoration={brace,amplitude=7pt, mirror}] (e.south west) -- (f.south east) node[midway, below=8pt] {morphological tags};

\draw[decorate,decoration={brace,amplitude=5pt, mirror}] (g.south west) -- (j.south east) node[midway, below=8pt] {form};

\end{tikzpicture}
}
\end{center}
\caption{The morphological inflection task: an input-output example, inflection of \en{} lemma ``well'' to superlative, inflected form ``best'' (figure adopted from \citet{tsd}).} 
\label{fig:inflection-task-input-output-formulation}
\end{figure}

\section{Introduction}

Morphological inflection, the task of producing the inflected form from a base word form (lemma) to express grammatical categories (see \cref{fig:inflection-task-input-output-formulation}), has become a highly active research area in recent years, largely driven by the \sig{} shared tasks held from 2016 to 2023 \citepSTs{}. 
The traditional approach to inflection has been to treat the lexical entries of lemma-tag-form triples uniformly, without any information about their frequency distribution. This is natural when a morphological lexicon (e.g., \um{} (UM) \citep{unimorph-40-batsuren-etal-2022}) is used, which is a prevalent approach.

However, in production deployment of a morphological inflection system, we may expect the user input to reflect a real-world distribution of frequencies: users will probably enter generally frequent words more commonly than rare words. This fact has been ignored for decades until quite recently \citep{kodner-reality-2023,tsd}.

With a focus on future deployment, we address this gap by exploring the incorporation of corpus frequency information into the task of morphological inflection along three key axes: train-dev-test split, data sampling during training, and evaluation, experimenting with Universal Dependencies (UD) \citep{ud-v2.14} corpora in 43 languages.

Our contributions are as follows:
\begin{enumerate}
    \item We experiment with a new train-dev-test split technique that satisfies both recent methodological standards (being lemma-disjoint \citep{goldman-2022-lemma-overlap, st23-goldman-etal-2023-sigmorphon}) and practical requirements for real-world deployment by ensuring a realistic distribution of items with varying corpus frequencies through frequency-weighted sampling \citep{kodner-reality-2023,tsd}.

    \item Throughout the experiments, we complement the standard \uacc{} evaluation with \wacc{} \citep{one-thousand-nicolai}, a previously proposed but never used metric that better reflects the deployment conditions by assigning greater importance to frequent words.

    \item Furthermore, motivated by the focus on frequent words in evaluation, we conduct pioneering experiments with \textit{frequency-aware} training; directly incorporating the importance (corpus occurrence counts) of training data items (\ltf{} triples) during training by frequency-weighted sampling. Our results indicate that this approach is promising for the majority of the languages.
\end{enumerate}

We make all our work open-source by publicly releasing our code at GitHub: \url{https://github.com/tomsouri/corpus-frequencies-in-inflection}.

\section{Related Work}

In the field of morphological inflection, there has been a lot of recent work, mostly pushed forward by the \sig{} shared tasks \citepSTs{}. The shared tasks leveraged the \um{} database \citep{unimorph-20-kirov-2018,unimorph-30-mccarthy-2020,unimorph-40-batsuren-etal-2022}, an ever-growing morphological lexicon. The data used in the shared tasks, as well as in other relevant works, are used in the form of a lexicon of unique \ltf{} triplets, without any notion of their frequency: the train, dev and test sets are simple lists of the triplets.

\subsection{Train-Dev-Test Data Split}
Until 2021, common practice was to split the lexicon of \ltf{} triplets into train, dev, and test uniformly randomly, without any additional control over the lemma overlap or feature overlap between train and test. \citet{wug-test-liu-hulden-2022-transformer} showed that lemma overlap artificially inflates the model's performance. \citet{goldman-2022-lemma-overlap} investigated this problem more and suggested a lemma-disjoint split technique, which was then adopted by the next installation of the \sig{} shared task \citep{st23-goldman-etal-2023-sigmorphon}.

Finally, \citet{kodner-reality-2023} have focused on different split practices and thoroughly evaluated and discussed their properties. In addition to the issues with overlap control, he also revealed drawbacks of standard \textit{ uniform sampling}. Uniform sampling over a set of triplets lemma-tag-form means that we sample them uniformly randomly into train, dev, and test, regardless of their corpus occurrence frequencies.\footnote{This is the only approach one may use if working with a morphological lexicon that only contains the \ltf{} triplets and lacks the information about their frequency in natural text.}
According to \citet{kodner-reality-2023}, this leads to an unnatural train-test split, with a bias towards low-frequency and thus more regular and reliable training items. To mitigate this problem, they suggest a new, \textit{frequency-weighted} split strategy: the types (\ltf{} triples) are partitioned at random weighted by their frequency in a corpus. The train set is sampled first, then the dev+test is sampled and then uniformly separated. This sampling biases the train set towards items with high corpus frequency, and dev and test towards low-frequency items, leading to a more realistic train-test distribution.
\citet{tsd} have built on this and proposed a new split technique that combines the frequency-weighted split with the lemma-disjoint approach. We employ this new technique, as the realistic train-test frequency distribution fits better our needs, and lemma-disjoint split allows evaluating generalization abilities.

\subsection{Evaluation}
When evaluating morphological inflection, exact-match accuracy is used, as the ratio of correctly predicted forms (also called \textit{type accuracy} or simply \textit{accuracy}, see \cref{uacc}), on a lexical test set, not considering the corpus frequencies of the items.

\begin{equation}
    \label{uacc}
    \text{type acc}=\dfrac{1}{\text{total items}}\cdot\sum_{\substack{\text{correct}\\\text{items}}}{1}
\end{equation}

\citet{one-thousand-nicolai} proposed to use the so called \textit{\wacc{}} (see \cref{wacc}), where the items are weighted by their corpus frequency. 

\begin{equation}
    \label{wacc}
    \text{token acc}=\dfrac{1}{\text{total occurrences}}\sum_{\substack{\text{correct}\\\text{items}}}{\text{occurrences(item)}}
\end{equation}

They argue that although \uacc{} is easy to compute from the morphological lexicons, it ``\textit{may over-represent rare forms, which are often regularly inflected and thus simpler to produce}'' \citep[Section 7]{one-thousand-nicolai}.
However, they conclude that computing \wacc{} is infeasible because it would require a corpus of running text annotated for morphological inflection, and they themselves do not evaluate using that metric, but only its approximation. 
To our knowledge, no further work on morphological inflection used the \wacc{} for the evaluation.

To meet the standards, we report traditional type accuracy and also include the proposed token accuracy, which better reflects real-world inflection system usage.

\subsection{Frequency-Aware Training}

To the best of our knowledge, we are not aware of previous relevant work in morphological inflection that incorporates corpus frequencies in (supervised) machine learning context, especially a neural-network-based one.

\section{Methodology}

With the focus on future deployment, we decide to perform the frequency-weighted train-dev-test split (for a realistic train-test distribution), evaluate with \wacc{}, which corresponds to evaluating on a running text and therefore reflects a real-world usage, and experiment with frequency-aware training.

\subsection{Data Selection}
\label{data-selection}

To be able to experiment with the techniques using corpus frequency information, we need to first have that information. The commonly used data source, \um{} (UM), is a lexicon and completely lacks information on frequency. 

We consider several possible approaches to obtain morphological data (\ltf{} triples) along with their corpus frequencies:

\begin{enumerate}
    \item Use a manually, morphologically annotated corpus of running text, such as Universal Dependencies (UD) \citep{ud2-nivre-etal-2020} directly as the data source. This approach has the advantage that it is simple and does not require any matching or tag conversion. Furthermore, if UD is used as the annotated corpus, the morphological tags are already in the CoNLL-U format, which is suitable for connecting the system into pipelines with other systems that work with UD (for example, UDPipe \citep{udpipe:2017}).
    However, using UD as the only source of morphologically annotated data could lead to lower coverage of lemmas/forms, compared to using \um{}.
    This approach was used by \citet{tsd}, and a variation of it (specifically intersecting an annotated corpus with \um{} and thus enriching \um{} with corpus frequencies) was used by \citet{kodner-reality-2023}.

    \item Use a raw corpus of running text and align it with UM, while dropping UM entries with corpus count = 0, that is, not covered by the corpus. This would lead to slightly higher coverage of lemmas/forms should a large corpus be used. However, for a proper alignment, we need to resolve ambiguities of two types: a form equal by string equality, corresponding to two different morphological categories of the same lemma (e.g., form ``\textit{hradu}'' as both dative/locative of lemma ``\textit{hrad}'' (\textit{castle} in Czech)), or even corresponding to two different lemmas (e.g., form ``\textit{ženu}'' as form of lemma ``\textit{žena}'' (\textit{woman} in Czech) or as form of lemma ``\textit{hnát}'' (\textit{to hurry} in Czech)). 
    For resolving such ambiguities, we consider several possible approaches:
    \begin{enumerate}
        \item divide counts evenly between ambiguous forms (used by \citet{kodner-reality-2023}), which is problematic, because in reality, the counts are uneven;
        \item use a tag-based back-off, either choosing the more frequent tag in general, no matter the lemma, or dividing the counts proportionally to the overall distribution of the tags, which is still not realistic, but it may work as a good approximation;
        \item use a tagger (a morphological analyzer, such as UDPipe \citep{udpipe:2017}) to automatically annotate the raw corpus, which is problematic because of high computational demands to run the tagger on large running texts, and also brings a potential leak of test data, as UDPipe was trained on a part of UD from which some lemmas are in our test set.
    \end{enumerate}
    Nevertheless, none of these techniques to resolve the ambiguities removes the problem of still having a high number of UM items with zero corpus occurrences and thus throwing away a lot of data.
    \item Use option 2) and do not drop UM items with 0 count: Since we work with corpus counts as numbers, we could assign some constant (less than 1) as the corpus count to all theoretically possible forms (those present in UM) that are not found in the corpus of running text.
\end{enumerate}

After carefully considering the advantages and drawbacks of each method, we select the first method, using the annotated corpus of running text (UD) directly as the data source, as it is the most straightforward approach, while keeping in mind that it leads to lower coverage of lemmas/forms, and we leave the other approaches for future work.

From Universal Dependencies, we select corpora for 43 languages,\footnote{See \cref{tab:test-joint} for the selected languages and corpora.}
specifically, \ie{} languages \writtenLatinScript{} and languages spoken in Europe. We lexicalize them by extracting unique \ltf{} triples along with their occurrence counts for training and evaluation. For hyperparameter tuning, we select a subset of five development languages: Czech, English, Spanish, Breton, and Basque.

\subsection{Train-Dev-Test Split}

The canonical split of UD is not suitable for the task of morphological inflection, because a naive extraction of \ltf{} triples from the canonical UD split naturally produces a train-test overlap, both on the item level (\ltf{} triples), and on the lemma level.
Therefore, we need to resplit the data. 
We aim for potential future deployment, therefore, we need a frequency-weighted split, as advocated by \citet{kodner-reality-2023}, for a realistic train-test frequency distribution. However, their proposed split is not lemma-disjoint and thus breaks recent methodological requirements \citep{goldman-2022-lemma-overlap,st23-goldman-etal-2023-sigmorphon} by obscuring an evaluation of the model's generalization abilities. To meet both requirements, we employ a new split technique proposed by \citet{tsd}, which combines the lemma-disjoint approach with the frequency-weighted approach. It consists of the following steps:
\begin{enumerate}
    \item 
    obtain the total occurrence count for each lemma in the corpus,
    \item sample lemmas first into the train set, randomly, weighted by the occurrence counts (until we obtain a desired amount of data in the train set),
    \item sample lemmas from the rest uniformly randomly (no weighted sampling) to the dev set (until a desired amount of data in the dev set is achieved),
    \item put the rest of lemmas into the test set.
\end{enumerate}

We use a train:dev:test ratio 8:1:1 in terms of total occurrence counts.

\subsection{Evaluation}
For evaluating the inflection systems, we use the standard \textit{\uacc{}} (also called simply \textit{accuracy}, see \cref{uacc}), working at the level of a lexicon, and complement it with \textit{\wacc{}} (see \cref{wacc}) \citep{one-thousand-nicolai}, which gives more weight to frequent words. 

In its natural interpretation, \uacc{} corresponds to evaluation on a lexicon of unique \ltf{} triples (ignoring frequencies), while \wacc{} corresponds to evaluation over a running text annotated with lemmas and tags (naturally reflecting frequencies). Therefore, \wacc{} better reflects real-world usage of an inflection system.
Clearly, \wacc{} is better suited for distinguishing between two models when one is superior in predicting more frequent words --- a situation that \uacc{} cannot detect.

\subsection{Model}
\label{sec:methodology-model}
For model training, we use a state-of-the-art, small-capacity, encoder-decoder, sequence-to-sequence \tr{} architecture, trained from scratch on the inflection data, with the lemma-tag pair as input and the inflected form as output (see \cref{fig:inflection-task-input-output-formulation}), as standard in the field.
We tune the model's hyperparameters on the five development languages (see \cref{tab:hyperparams-tuning} for the final values).
The tuned model (without frequency-aware training) outperforms both \sig{} neural baselines \citep{wu-etal-2021-applying-first} in all five development languages.

\begin{table}[]
    \centering
    \begin{tabular}{l l @{\hspace{7em}} l l @{\hspace{7em}} l l}
    \toprule
    hyperparam & value & hyperparam & value & hyperparam & value \\
    \midrule
    \adrop{}     & 0.1   & \lnorm{}    & 0.01  & \bs{}      & 512   \\
    \ddrop{}     & 0.15  & \head{}     & 4     & \lr{}      & 0.001 \\
    \fdrop{}     & 0.35  & \layers{}   & 3     & \maxeps{}  & 960   \\
    \ldrop{}     & 0.2   & \layerdim{} & 256   & \ckptsel{} & on    \\
    \clipnorm{}  & 1.0   & \ffnndim{}  & 64    &  \decay{}  & cosine  \\
    \bottomrule
    \end{tabular}

    \caption[Hyperparameters - setup]{Hyperparameters as tuned on the five development languages and used for all experiments. The model is trained with Adam. The checkpoint is selected based on dev performance (token accuracy) for each language separately.}
    \label{tab:hyperparams-tuning}
\end{table}

\subsection{Frequency-Aware Training}
As we give more weight to frequent words in evaluation with \wacc{}, one might ask whether it could be beneficial to give more weight to such words also in training. 
To support this, we introduce sample weights and use weighted random sampling to sample training data items into batches.

If we wanted to mimic training on a running corpus, the direct approach would be to use raw frequency counts of the data items as \sweights{}.
Another specific case is setting all \sweights{} to 1, which would lead to uniform sampling, which is a standard approach in morphological inflection.
For a continuous scale between these two extremes, we introduce a \textit{\witemp{}}, denoted as \witletter{}, a real-valued hyperparameter, usually $\witletterM{}\in[0,1]$. 
For training data $T$, denote by $c_T(A)$ the corpus occurrence count for a data item (\ltf{} triple) $A$. Let $M_T$ be the sum of $c_T(B)$ on all \ltf{} triples $B$ (the total number of \ltf{} triples in training data before aggregation, which can be viewed as the original corpus length in tokens) and let $V_T$ be the number of unique \ltf{} triples in training data. The \sweight{} for weighted random sampling is computed from the raw corpus count and the \witemp{} value \witletter{}:

\begin{equation}
    w_T(A)=\left(c_T(A)\right)^\witletterM{}
    \label{eq:wit-sample-weights}
\end{equation}

The probability that the training data item $A$ will be sampled in an individual sampling step is given by the \sweight{} with normalization:

\begin{equation} 
    p(A)=\dfrac{\left(c_T(A)\right)^\witletterM{}}{\sum_{B}(c_T(B))^\witletterM{}}
    \label{eq:wit-prob-of-sampling-A} 
\end{equation}

The ratio of the probabilities of two data items $A$ and $B$ being sampled in a single sampling step is:

\begin{equation} 
\frac{p(A)}{p(B)} = \left(\dfrac{c_T(A)}{c_T(B)}\right)^\witletterM{}
\label{eq:ratio-A-B} 
\end{equation}

\subsubsection{Specific Values of Corpus-Frequency Temperature and Their Interpretation}

With $\witletterM{}=1$, \sweights{} are equal to the raw corpus frequencies, and the probability that the training data item $A$ will be sampled in an individual sampling step is then
\begin{equation}
p_{\witletterM{}=1}(A)=\dfrac{c_T(A)}{\sum_{B}c_T(B)}=\dfrac{c_T(A)}{M_T}
\label{eq:wit-temp-is-1}
\end{equation}

The value $\witletterM{}=0$ corresponds to the uniform \sweights{}, and the probability that any specific data item will be sampled in an individual sampling step is then 
\begin{equation}
p_{\witletterM{}=0}(A)=\dfrac{1}{V_T}    
\label{eq:wit-temp-is-0}
\end{equation}

With $\witletterM{}=1/2$, \sweights{} are equal to the square root of raw corpus frequencies.
Then item $A$ with raw occurrence count 400 would be sampled with $\sqrt{400}=20$ times higher probability than item $B$ with raw occurrence count 1 (with \sweight{} $\sqrt{1}=1$).

\Witemp{} values from the interval $[0,1]$ correspond to the continuous scale between the two extremes (uniform \sweights{} ignoring the corpus frequencies with $\witletterM{}=0$, and \sweights{} equal to the raw corpus frequencies $\witletterM{}=1$).
Nevertheless, it is technically possible to set it also to values outside the range, and it also has a natural meaning. 

The value $\witletterM{}=-1$ leads to \sweights{} being the inverse of the raw corpus counts. Then the ratio of the probabilities of two data items $A$, $B$ being sampled in a single sampling step is
\begin{equation} 
\frac{p_{\witletterM{}=-1}(A)}{p_{\witletterM{}=-1}(B)} = \dfrac{c_T(B)}{c_T(A)}
\label{eq:ratio-A-B-with-temp--1} 
\end{equation}
That is, less frequent data items are sampled more frequently during training. The value $\witletterM{}=2$ leads to extremely emphasizing the frequent data items. In the example of $c_T(A)=400$, $c_T(B)=1$, with $\witletterM{}=2$, $A$ would be sampled 160,000 times more frequently than $B$ during training.

\section{Experiments and Results}

To measure the effect of \wit{}, we experiment with the tuned \tr{} model (as described in \cref{sec:methodology-model}), trained for 960 epochs with checkpoint selection based on \wacc{} on the dev set, with different values of \witemp{}. Our hypothesis is that giving more importance to frequent words during training could improve the performance of the system when evaluating with \wacc{}, which itself gives more importance to frequent words in evaluation.

We experiment with different values of \witemp{} (\witletter{}) from the range $[0,1]$. For experimental purposes, we also try \witletter{} values 1.1 and 2, which lead to over-emphasizing the frequent items during training, and negative values of \witletter{}, which go directly against our objective (giving more importance to frequent words), as they lead to focusing more on rare words during training.

\subsection{Effect of Corpus-Frequency Temperature Value \witletter{} on Token Accuracy}

\begin{table}[htbp]

\caption{Dev results, reported \textbf{\wacc{}} (\%) for different \witemp{} values (\witletter{}). Macro average over all languages is reported in the rightmost column in the bottom part of the table. A separate color scale was applied to each language, shading from dark red (lowest) through white (median) to dark green (highest).}
\label{tab:dev-wacc}
\centering

\resizebox{0.93\hsize}{!}{%
\begin{tabular}{rcccccccccccc}%
\toprule

\witletter{} & Afrikaans & Basque & Belarusian & Breton & Bulgarian & Catalan & Croatian & Czech & Danish & Dutch & English \\
-1.0 & \cellcolor{red!47!white} 85.61 & \cellcolor{red!78!white} 87.51 & \cellcolor{red!78!white} 88.14 & \cellcolor{mygreen!26!white} 70.40 & \cellcolor{red!63!white} 92.67 & \cellcolor{red!68!white} 95.60 & \cellcolor{red!78!white} 92.65 & \cellcolor{red!57!white} 97.67 & \cellcolor{red!36!white} 90.18 & \cellcolor{red!68!white} 78.36 & \cellcolor{mygreen!68!white} 96.34 \\
-0.8 & \cellcolor{red!57!white} 85.54 & \cellcolor{red!68!white} 87.89 & \cellcolor{red!68!white} 88.34 & \cellcolor{red!26!white} 70.13 & \cellcolor{red!63!white} 92.67 & \cellcolor{red!57!white} 95.81 & \cellcolor{red!68!white} 92.82 & \cellcolor{red!36!white} 97.72 & \cellcolor{mygreen!26!white} 90.39 & \cellcolor{red!57!white} 78.97 & \cellcolor{mygreen!84!white} 96.38 \\
-0.6 & \cellcolor{red!36!white} 85.65 & \cellcolor{red!57!white} 88.20 & \cellcolor{red!57!white} 88.71 & \cellcolor{red!36!white} 70.10 & \cellcolor{red!36!white} 92.72 & \cellcolor{red!47!white} 95.94 & \cellcolor{red!57!white} 92.90 & \cellcolor{mygreen!15!white} 97.81 & \cellcolor{mygreen!36!white} 90.44 & \cellcolor{red!47!white} 79.34 & \cellcolor{red!15!white} 96.19 \\
-0.5 & \cellcolor{red!26!white} 85.72 & \cellcolor{red!47!white} 88.47 & \cellcolor{red!47!white} 88.81 & \cellcolor{mygreen!57!white} 70.67 & \cellcolor{red!26!white} 92.81 & \cellcolor{red!26!white} 96.19 & \cellcolor{red!42!white} 93.06 & \cellcolor{red!15!white} 97.78 & \cellcolor{red!26!white} 90.32 & \cellcolor{red!36!white} 79.65 & \cellcolor{mygreen!100!white} 96.39 \\
\bottomrule

\end{tabular}
}
\end{table}

In \cref{tab:dev-wacc}, we report the performance comparison in terms of \wacc{} on the dev set with different \witemp{} values for each language separately.

We observe that $\witletterM{}=2.0$ leads to a complete failure,\footnote{With the only exception of Old French, where $\witletterM{}=2.0$ leads to the best performance.} probably because of the extreme emphasis on the frequent words.
When we focus on $\witletterM{}\in{}[-1,1]$, we can observe a clear tendency in some languages (Basque, Old French, Turkish): the higher the temperature value, the better the performance. In a larger set of languages we can see more or less consistent improvement when increasing $\witletterM{}$ from -1 to 0.5 (or another value close 0.5), followed by a drop, again more or less consistent: Belarusian, Bulgarian, Croatian, Dutch, German, Gothic, Irish, Italian, Latin, Polish, Pomak, Portuguese, Russian, Spanish, and Welsh.
In other languages, \wit{} does not seem to have a major effect, as the best performance is achieved with temperature around 0 (uniform sampling): Afrikaans, Catalan, Czech, Danish, Estonian, Finnish, French, Greek, Hungarian, Icelandic, Latvian, Manx, Norwegian, Romanian, Scottish Gaelic, Slovenian and Swedish.

Quite interestingly, there are also languages in which negative values of \witemp{} (which give more weight to rare forms during training) lead to the best performance, such as English, Galician, and Slovak.
In other languages, such as Breton, Lithuanian, Low Saxon, or Ukrainian, the performance at different temperature levels is quite noisy and does not provide any insight.

On average in all languages (rightmost column in the bottom part of \cref{tab:dev-wacc}), there is a continuous improvement when the temperature increases to 0.5, and then a continuous decline, indicating that if we shall select a single temperature value to be used in all languages, it would be $\witletterM{}=0.5$ (sample weights equal to the square root of the raw corpus occurrence counts).

To sum up, for almost half of the languages the performance corresponds to our hypothesis that giving more importance to frequent words during training should increase the performance of the system in terms of \wacc{}: either steadily with the best results achieved with $\witletterM{}=1$,  weights being the raw corpus occurrence counts (Basque, Old French, Turkish), or rather up to the square root of occurrence counts ($\witletterM{}=0.5$).
On the other hand, there are languages where \wit{} does not seem to have a major effect, and even some with opposite results (English, Galician, Slovak).

\subsection{Effect of Corpus-Frequency Temperature Value \witletter{} on Type Accuracy}

\begin{table}[htbp]
\caption{Dev results, reported \textbf{\uacc{}} (\%) for different \witemp{} values (\witletter{}). Macro average over all languages is reported in the rightmost column in the bottom part of the table. A separate color scale was applied to each language, shading from dark red (lowest) through white (median) to dark green (highest).}
\label{tab:dev-uacc}
\centering

\resizebox{0.93\hsize}{!}{%
\begin{tabular}{rcccccccccccc}%
\toprule

\witletter{} & Afrikaans & Basque & Belarusian & Breton & Bulgarian & Catalan & Croatian & Czech & Danish & Dutch & English \\
-1.0 & \cellcolor{red!57!white} 82.82 & \cellcolor{red!78!white} 86.37 & \cellcolor{red!78!white} 87.47 & \cellcolor{red!15!white} 64.41 & \cellcolor{red!42!white} 91.13 & \cellcolor{red!47!white} 93.48 & \cellcolor{red!57!white} 91.95 & \cellcolor{red!47!white} 97.35 & \cellcolor{red!36!white} 89.44 & \cellcolor{red!57!white} 75.71 & \cellcolor{mygreen!36!white} 93.44 \\
-0.8 & \cellcolor{red!47!white} 82.85 & \cellcolor{red!68!white} 86.66 & \cellcolor{red!68!white} 87.65 & \cellcolor{red!31!white} 64.27 & \cellcolor{red!57!white} 91.08 & \cellcolor{red!36!white} 93.56 & \cellcolor{red!68!white} 91.87 & \cellcolor{red!31!white} 97.40 & \cellcolor{mygreen!47!white} 89.75 & \cellcolor{red!47!white} 76.09 & \cellcolor{mygreen!52!white} 93.47 \\
-0.6 & \cellcolor{red!36!white} 82.86 & \cellcolor{red!57!white} 86.81 & \cellcolor{red!57!white} 87.92 & \cellcolor{red!47!white} 64.22 & \cellcolor{red!26!white} 91.17 & \cellcolor{red!26!white} 93.68 & \cellcolor{red!47!white} 92.04 & \cellcolor{mygreen!21!white} 97.45 & \cellcolor{mygreen!36!white} 89.70 & \cellcolor{red!36!white} 76.39 & \cellcolor{red!15!white} 93.25 \\
-0.5 & \cellcolor{red!15!white} 83.01 & \cellcolor{red!36!white} 87.22 & \cellcolor{red!47!white} 87.96 & \cellcolor{mygreen!15!white} 64.61 & \cellcolor{red!42!white} 91.13 & \cellcolor{mygreen!15!white} 93.79 & \cellcolor{red!26!white} 92.14 & \cellcolor{mygreen!5!white} 97.44 & \cellcolor{red!15!white} 89.52 & \cellcolor{red!26!white} 76.63 & \cellcolor{mygreen!100!white} 93.65 \\
\bottomrule

\end{tabular}
}
\end{table}

In \cref{tab:dev-uacc}, we report a similar comparison, now evaluated with \uacc{} (which disregards the corpus occurrence counts of the evaluation items, treating all items uniformly). Although the results appear to be slightly more noisy than with \wacc{}, we observe similar trends in most of the languages: the continuous improvement in Basque and Turkish when increasing the temperature value to 1.0, and the languages with a peak close to $\witletterM{}=0.5$ (Belarusian, Dutch, German, Irish, Italian, Latin, Low Saxon, Portuguese, Scottish Gaelic, and Welsh). This is quite surprising: why should we benefit from giving more importance to frequent words in training if we disregard the frequencies in evaluation? We discuss it further in \cref{sec:discussion}.

\subsection{Token vs. Type Accuracy}

Another interesting observation is that although \wacc{} and \uacc{} differ in the absolute value (compare English with $\witletterM{}=0.0$ with 96.22\% in \wacc{} and 93.37\% in \uacc{}), they seem to be similar in terms of model ranking, at least when comparing systems trained with the same hyperparameters with the only difference in temperature value (and it is a trend we observed also during the development experiments with hyperparameter tuning).

To further investigate whether using \uacc{} as the main objective during the experiments could actually affect the final performance in \wacc{} (our primary objective for deployment of the system), we conduct the following experiment: we first train a system, selecting the checkpoint based on \wacc{}, and evaluate using \wacc{}. Then, we train another system, selecting the checkpoint based on \uacc{}, and again evaluate using \wacc{}. If a significant drop in performance is observed, it would suggest that relying on \uacc{} for design decisions during experimentation might lead to a worse final performance in \wacc{}.

For different maximum numbers of training epochs (60, 120, 240, 480, and 960), we run five instances\footnote{Differing by the random seed used for initialization.} of the proposed experiment on the five development languages. We find that performance drops are negligible: in \eus{}, the largest absolute drop is 0.7\%; in \es{} and \en{}, it is 0.3\%; in \cs{}, 0.2\%; and in \br{}, 0.9\%.

The fact that the differences are minor may be attributed to the nature of our splits --- data with high corpus-frequency counts tend to appear in the training set, resulting in a limited variance in the development set. The difference between \wacc{} and \uacc{} should be explored further, as we have not shown a substantial difference between them in terms of the relative ordering of models under our split and evaluation conditions.

\subsection{Test Comparison}
For the evaluation on the test set, we use four different systems: a naive \copybs{} baseline, which copies the input lemma to output, a model without \wit{} (with uniform sampling to batches, ignoring the corpus frequencies, denoted by $\witletterM{}=0.0$), a model with \wit{} (weighted random sampling into batches during training) with $\witletterM{}=0.5$ (the best value in average over all languages, weights being the square root of the raw corpus occurrence count), and finally, \witbest{}, a model that for each language separately selects the best temperature value based on the performance on the dev set (\wacc{}) and uses \wit{} with the selected temperature value.

\begin{table}[htbp]
    \centering
    \caption{Test results, \textbf{\wacc{}} (left) and \textbf{\uacc{}} (right). The best system for each language (selected separately for \wacc{} and \uacc{}) is marked with \textbf{bold}. \copybs{} copies the input lemma to output. $\witletterM{}=0.0$ is uniform sampling. $\witletterM{}=0.5$ is \wit{} with weights equal to the square root of the raw occurrence counts. \witbest{} is system, where the specific \witletter{} value is selected separately for each language, based on \wacc{} on dev set.}
    \label{tab:test-joint}
\begin{tabular}{l l
S[table-format=2.2] S[table-format=2.2] S[table-format=2.2] S[table-format=2.2] 
c 
S[table-format=2.2] S[table-format=2.2] S[table-format=2.2] S[table-format=2.2]
}
\toprule
lang $\downarrow$ & corpus  $\downarrow$ & \multicolumn{4}{c}{\wacc{} (\%)} & & \multicolumn{4}{c}{\uacc{} (\%)} \\

\cmidrule(lr){1-2} \cmidrule(lr){3-6} \cmidrule(lr){8-11}

system $\rightarrow$ &  & \copybs{} & \multicolumn{1}{c}{$\witletterM{}=0.0$} & \multicolumn{1}{c}{$\witletterM{}=0.5$} & \multicolumn{1}{c}{\witbest{}} & & \copybs{} & \multicolumn{1}{c}{$\witletterM{}=0.0$} & \multicolumn{1}{c}{$\witletterM{}=0.5$} & \multicolumn{1}{c}{\witbest{}} \\

\cmidrule(lr){1-2} \cmidrule(lr){3-6} \cmidrule(lr){8-11}

Afrikaans  & AfriBooms &  72.19 & 85.89 & \textbf{86.41} & 86.31  & & 69.18 & 82.61 & 82.93 & \textbf{83.82} \\
Basque  & BDT &  45.23 & 88.62 & 89.02 & \textbf{89.59}  & & 39.74 & 85.71 & 85.79 & \textbf{87.57} \\
Belarusian  & HSE &  43.44 & 88.64 & \textbf{88.90} & \textbf{88.90}  & & 40.88 & 87.07 & 87.38 & \textbf{88.14} \\
Breton  & KEB &  61.54 & 67.63 & 67.40 & \textbf{69.06}  & & 57.66 & 63.56 & 63.55 & \textbf{65.04} \\
Bulgarian  & BTB &  45.08 & \textbf{92.64} & 92.40 & 92.41  & & 37.90 & 90.61 & 90.81 & \textbf{91.06} \\
Catalan  & AnCora &  73.01 & 96.06 & \textbf{96.12} & 96.11  & & 64.73 & 93.18 & 93.35 & \textbf{93.49} \\
Croatian  & SET &  40.50 & 93.05 & 93.11 & \textbf{93.25}  & & 36.10 & 91.64 & 91.80 & \textbf{92.29} \\
Czech  & PDT &  45.96 & 97.41 & 97.48 & \textbf{97.52}  & & 37.53 & 97.38 & 97.43 & \textbf{97.50} \\
Danish  & DDT &  59.24 & \textbf{90.54} & 90.31 & 90.49  & & 55.61 & 89.84 & 89.53 & \textbf{89.87} \\
Dutch  & Alpino &  61.60 & 80.71 & \textbf{81.76} & \textbf{81.76}  & & 55.48 & 76.40 & 76.36 & \textbf{78.13} \\
English  & EWT &  82.10 & \textbf{96.11} & 95.84 & 96.03  & & 76.67 & 93.77 & 93.70 & \textbf{93.91} \\
Estonian  & EDT &  30.33 & 90.52 & 90.70 & \textbf{90.71}  & & 23.24 & 88.74 & 88.97 & \textbf{89.53} \\
Finnish  & TDT &  30.27 & \textbf{91.83} & \textbf{91.83} & \textbf{91.83}  & & 23.88 & 90.47 & 90.30 & \textbf{90.69} \\
French  & GSD &  74.27 & 96.71 & \textbf{96.79} & 96.74  & & 72.55 & 95.32 & 95.21 & \textbf{95.44} \\
Galician  & TreeGal &  63.39 & 92.61 & 92.21 & \textbf{92.75}  & & 59.67 & 93.86 & \textbf{94.01} & \textbf{94.01} \\
German  & GSD &  75.35 & 89.61 & \textbf{89.92} & \textbf{89.92}  & & 75.22 & 88.77 & 88.97 & \textbf{89.34} \\
Gothic  & PROIEL &  30.62 & 75.89 & 77.26 & \textbf{77.72}  & & 17.50 & 69.71 & 70.63 & \textbf{72.45} \\
Greek  & GDT &  40.62 & \textbf{84.70} & 84.52 & \textbf{84.70}  & & 34.45 & 81.63 & 81.41 & \textbf{82.04} \\
Hungarian  & Szeged &  56.17 & \textbf{93.50} & 93.37 & 93.44  & & 51.34 & 92.52 & \textbf{92.84} & 92.64 \\
Icelandic  & Modern &  41.48 & \textbf{72.36} & 72.24 & \textbf{72.36}  & & 35.71 & 68.34 & 69.10 & \textbf{70.02} \\
Irish  & IDT &  55.69 & 86.60 & \textbf{87.40} & \textbf{87.40}  & & 51.93 & 83.40 & 83.78 & \textbf{84.17} \\
Italian  & ISDT &  68.49 & 95.74 & 95.86 & \textbf{96.04}  & & 62.95 & 93.82 & 94.04 & \textbf{95.12} \\
Latin  & ITTB &  28.87 & 81.29 & \textbf{83.08} & 80.99  & & 14.58 & 74.25 & 74.74 & \textbf{77.39} \\
Latvian  & LVTB &  33.59 & \textbf{96.81} & 96.73 & 96.79  & & 27.81 & 96.32 & 96.41 & \textbf{96.42} \\
Lithuanian  & ALKSNIS &  31.20 & 93.93 & 93.92 & \textbf{93.96}  & & 27.79 & 92.51 & \textbf{92.77} & 92.71 \\
Low\ Saxon  & LSDC &  \textbf{59.03} & 55.03 & 56.49 & 56.32  & & \textbf{54.42} & 50.84 & 51.23 & 51.55 \\
Manx  & Cadhan &  66.20 & 71.47 & \textbf{71.69} & 71.47  & & 65.15 & 63.97 & 65.09 & \textbf{66.41} \\
Norwegian  & Bokmaal &  58.60 & \textbf{93.40} & 93.34 & 93.07  & & 53.62 & 92.68 & 92.80 & \textbf{93.00} \\
Old\ French  & PROFITEROLE &  \textbf{46.75} & 0.71 & 3.70 & 10.80  & & \textbf{19.82} & 0.10 & 0.11 & 0.31 \\
Polish  & PDB &  30.53 & 93.93 & 94.03 & \textbf{94.17}  & & 28.42 & 93.69 & \textbf{93.88} & 93.86 \\
Pomak  & Philotis &  27.75 & \textbf{51.17} & 50.63 & 50.54  & & 20.98 & 42.19 & 43.69 & \textbf{45.27} \\
Portuguese  & Bosque &  72.11 & \textbf{96.93} & 96.91 & 96.91  & & 67.66 & 95.83 & 96.05 & \textbf{96.19} \\
Romanian  & RRT &  44.17 & 92.25 & \textbf{92.49} & 92.25  & & 42.00 & \textbf{92.46} & 92.38 & 92.31 \\
Russian  & SynTagRus &  33.01 & 94.64 & \textbf{94.87} & \textbf{94.87}  & & 26.35 & 92.76 & 92.94 & \textbf{93.28} \\
Sanskrit  & Vedic &  12.51 & 78.39 & 78.98 & \textbf{79.25}  & & 9.43 & 74.65 & 74.88 & \textbf{76.07} \\
Scottish\ Gaelic  & ARCOSG &  66.16 & 65.80 & \textbf{66.49} & 66.20  & & 59.71 & 59.00 & 59.41 & \textbf{61.39} \\
Slovak  & SNK &  35.75 & \textbf{94.57} & 94.56 & 94.33  & & 31.83 & 93.98 & \textbf{94.17} & 94.05 \\
Slovenian  & SSJ &  38.18 & 95.65 & 95.60 & \textbf{95.71}  & & 33.31 & 95.17 & 95.22 & \textbf{95.39} \\
Spanish  & AnCora &  70.42 & \textbf{97.31} & 97.30 & 97.22  & & 61.38 & 94.86 & 94.97 & \textbf{95.17} \\
Swedish  & Talbanken &  52.61 & 90.73 & 90.14 & \textbf{90.91}  & & 48.63 & 89.44 & 89.55 & \textbf{90.03} \\
Turkish  & BOUN &  51.91 & 84.46 & 84.66 & \textbf{85.16}  & & 42.80 & 80.70 & 81.10 & \textbf{81.97} \\
Ukrainian  & IU &  35.41 & \textbf{92.60} & 92.49 & 92.49  & & 32.11 & 92.51 & \textbf{92.55} & 92.51 \\
Welsh  & CCG &  65.16 & 88.15 & 88.15 & \textbf{88.27}  & & 60.11 & 84.39 & 84.41 & \textbf{84.95} \\

\cmidrule(lr){1-2} \cmidrule(lr){3-6} \cmidrule(lr){8-11}

\textbf{macro avg}  &  &  50.15 & 85.04 & 85.28 & \textbf{85.51}  & & 44.37 & 82.57 & 82.80 & \textbf{83.41} \\

\bottomrule
\end{tabular}
    
\end{table}

In terms of \wacc{} (see \cref{tab:test-joint}, left part), \wit{} with \witbest{} outperforms uniform sampling in 26 languages and achieves the overall best result in 23 languages out of 43 languages. It is outperformed by $\witletterM{}=0.5$ in 12 languages and by uniform sampling in additional 5 languages. That is relatively disappointing, and it probably shows overfitting to the dev set (by hyperparameter tuning, checkpoint selection and temperature value selection), because otherwise we would expect the \witbest{} model to outperform the rest of models.
Compared to uniform sampling, \witsqrt{} brings improvement in 23 languages and outperforms all other models in 13 languages. In two languages (Low Saxon and Old French), none of our models outperformed the  \copybs{} baseline.

In terms of \uacc{} (see \cref{tab:test-joint}, right part), the results are more straightforward: \witbest{} improves over uniform sampling in 41 out of 43 languages, achieving the overall best result in 35 languages. It is outperformed by \witsqrt{} in only 6 languages. As in \wacc{}, it is outperformed by \copybs{} baseline in two languages.
It is surprising that when choosing the temperature value based on \wacc{} performance on the dev set, on the test set it helps more the \uacc{} than the \wacc{}. 
It would be beneficial to explore this more in future work and to seek an explanation.

Regarding the comparison of \wacc{} and \uacc{}, note that at least in the test results we can observe some differences (\witbest{} system clearly better than the rest according to \uacc{}, yet not so clearly according to \wacc{}).

\section{Discussion}
\label{sec:discussion}

We found that, in at least some languages, \wit{} provides a clear benefit for both \wacc{} and \uacc{}.

If the data split was not lemma-disjoint, the reason for its positive effect on \wacc{} would be fairly straightforward: Suppose a frequent lemma A appears in the training set. Under \wit{}, it receives a higher weight, allowing the model to learn it more effectively. If the same lemma A also occurs in the development or test set, the model will (likely) inflect it correctly. Since lemma A is frequent, its correct inflection contributes disproportionately to the overall \wacc{}, due to the higher evaluation weight it receives.

Nevertheless, the split is lemma-disjoint, which raises a question: Why does \wit{} help? Furthermore, why should it help even in terms of \uacc{}? Why does it help only in some languages, and why we even achieve an opposite effect some other languages?

One possible explanation for why \wit{} is beneficial, even with the lemma-disjoint split, is that frequent words in the dev/test sets tend to inflect in a similar way to frequent words in the training set. If this is the case, assigning greater weight to frequent words during training should help the model inflect frequent dev/test words more accurately, thereby increasing \wacc{}. A more straightforward explanation is that \wit{} helps due to a derivational leak. For instance, a frequent lemma in the training set, such as the Czech verb ``\textit{jít}'' (``\textit{to go}''), may have derived lemmas in the dev/test sets, e.g., ``\textit{přijít}'' (``\textit{to come}''), ``\textit{projít}'' (``\textit{to go through}''). These derived forms are also frequent and thus receive a higher weight during evaluation. Since derived verbs in Czech inflect in very similar ways, giving more weight to ``\textit{jít}'' during training improves performance on ``\textit{přijít}'' and ``\textit{projít}'' during evaluation.

As for why \wit{} might improve also \uacc{}, the gain may come from the fact that frequent lemmas occur in corpus not only with a higher occurrence count (reflected in \wacc{}) but also with a greater variety of distinct forms. Since \uacc{} treats each \ltf{} triple equally, this increased form diversity could contribute to the improvement.

\subsection{Future Work}

Regarding why \wit{} helps in some languages but has the opposite effect in others, its success appears to depend on factors that are not yet well understood. Investigating these in future work would be worthwhile. Possible explanations include linguistic properties such as morphological richness and regularity, or corpus-related factors such as data size and variability.

In addition, other ways of obtaining corpus frequencies for morphological data could be explored, as described in \cref{data-selection}, especially using large raw data to extract the frequencies with some tag-based back-off for disambiguation, not dropping the UniMorph items with zero actual corpus occurrences.

\citet{kodner-reality-2023} argues that frequency-weighted split better reflects real-world frequency distribution of train-test, and so we use it.
However, it would be beneficial to explore more the split compared to uniform split, to experimentally answer whether it would hurt the real-world performance, if we developed a system on a uniform train-dev-test split.

Also, it would be worth experimenting with the canonical UD train-dev-test split (the original split of UD, not lemma-disjoint).

In terms of metrics (\wacc{} and \uacc{}), it would be worth further exploring whether there is a difference in terms of what metric we use during the development of a model, in contrast to the metric used in the final evaluation (that is, tuning hyperparameters for two systems, each according to one of the metrics, and then evaluate both systems according to \wacc{} and see if the performance differs). Furthermore, it would be particularly interesting to measure the differences between them on a uniform (not frequency-weighted) split of UD. Moreover, enriching a standard benchmark like \sig{} with corpus frequency information and re-evaluating the shared task using our \wacc{} metric could potentially yield valuable insights.

\section{Conclusion}

We explored the incorporation of corpus frequencies in the task of morphological inflection along three key dimensions: train-dev-test split of data (frequency-weighted lemma-disjoint split), training data sampling (\wit{}), and evaluation.
We designed and consistently applied a previously suggested but never before used evaluation metric that assigns greater importance to frequent word forms (\wacc{}).
We demonstrated that in our setting (lemma-disjoint, frequency-weighted split), the difference between the original metric (\uacc{} or just accuracy) and \wacc{} is minimal with respect to the relative ranking of different systems.
Our pioneering experiments with \wit{} showed promising results, improving over uniform sampling in 26 out of 43 languages. The overall best weighting scheme seems to be with corpus frequency temperature 0.5 (each sample is given a weight corresponding to the square root of its corpus frequency).

All three dimensions (frequency-weighted split, \wit{}, and \wacc{}) are worth further investigation.

\begin{acknowledgments}

This research was supported by the Johannes Amos Comenius Programme (P JAC) project No. CZ.02.01.01/00/22\_008/0004605, Natural and anthropogenic georisks.

Computational resources for this work were provided by the e-INFRA CZ project (ID:90254), supported by the Ministry of Education, Youth and Sports of the Czech Republic.

The work described herein uses resources hosted by the LINDAT/CLARIAH-CZ Research Infrastructure (projects LM2018101 and LM2023062, supported by the Ministry of Education, Youth and Sports of the Czech Republic).

We thank the anonymous reviewers for their valuable comments.

\end{acknowledgments}

\section*{Declaration on Generative AI}

During the preparation of this work, the authors used GPT-4 and Writefull to check grammar and spelling. After using these tools/services, the authors reviewed and edited the content as needed and take full responsibility for the publication’s content.

\bibliography{referencies}

\end{document}